\documentclass{article}
\usepackage{spconf,amsmath,graphicx}
\usepackage{epstopdf}
\usepackage{amssymb}
\usepackage{amsthm}
\usepackage{array}
\usepackage{array}
\newcommand{\PreserveBackslash}[1]{\let\temp=\\#1\let\\=\temp}
\newcolumntype{C}[1]{>{\PreserveBackslash\centering}p{#1}}
\newcolumntype{R}[1]{>{\PreserveBackslash\raggedleft}p{#1}}
\newcolumntype{L}[1]{>{\PreserveBackslash\raggedright}p{#1}}
\usepackage{amsmath}
\usepackage{type1cm}
\usepackage{slashbox}
\usepackage{bm}
\usepackage{multirow}

\usepackage{makeidx}         
\usepackage{graphicx}        

\usepackage{multicol}        
\usepackage[bottom]{footmisc}
\usepackage{subfigure}
\usepackage{booktabs}
\usepackage{algorithm, algorithmic} 

\usepackage{hyperref}

\title{Augmented Space Linear Model}
\name{Zhengda Qin$ ^1 $, Badong Chen$ ^1 $, Nanning Zheng$ ^1 $ and Jose C. Principe$ ^{1,2} $\thanks{Thanks to XYZ agency for funding.}}
\address{$^1$ Institute of Artificial Intelligence and Robotics, Xi'an Jiaotong University, Xi'an, 710049, China.
	\\$^2$  Computational NeuroEngineering Laboratory, University of Florida, Gainesville, FL 32611, USA. }
\begin{document}
%
\maketitle
\begin{abstract}
The linear model uses the space defined by the input to project the target or desired signal and find the optimal set of model parameters. When the problem is nonlinear, the adaption requires nonlinear models for good performance, but it becomes slower and more cumbersome. In this paper, we propose a linear model called \textit{Augmented Space Linear Model} (ASLM), which uses the full joint space of input and desired signal as the projection space and approaches the performance of nonlinear models. This new algorithm takes advantage of the linear solution, and corrects the estimate for the current testing phase input with the error assigned to the input space neighborhood in the training phase. This algorithm can solve the nonlinear problem with the computational efficiency of linear methods, which can be regarded as a trade off between accuracy and computational complexity. Making full use of the training data, the proposed augmented space model may provide a new way to improve many modeling tasks.
\end{abstract}
\section{Introduction}
\label{sec:intro}

Since the early work of Legendre and Gauss in the late XVIII century, linear or nonlinear regression has employed the space defined by the input data to project the target or desired response and find, in a training set, the optimal set of model parameters through mean square error minimization. This approach has been totally embraced by the adaptive signal processing \cite{haykin2008adaptive}, control theory, pattern recognition and machine learning communities \cite{duda2012pattern}, and has become the de facto standard to perform function approximation. 

The pursuit of this alternative is based on theoretic reasons, i.e. to expand the horizon of function approximation theory, but its impact on current machine learning applications is perhaps even higher. In the conventional modeling approach, when the system that created the input-desired data pairs is nonlinear, the linear model must be substituted by a nonlinear model (e.g. artificial neural networks), which means that optimization becomes nonlinear in the parameters. This implies that local minima exist in the performance surface, and gradient search techniques become slow, cumbersome and there is no guarantee of finding the optimal solution. This is one of the current bottlenecks of nonlinear modeling and machine learning. All these methods, meanwhile, ignore the error after training the parameter, but the available error information can be better utilized to provide a novel approach to function approximation, as we will demonstrate here.

Our vision is to create universal learning systems that are easy to train and guarantee to converge to the optimum, which we called \textit{convex universal learning machines} (CULMs) \cite{principe2015universal}. CULMs are universal mappers with architectures that either do not have a hidden layer or do not need to train the hidden layer weights. One distinctive class of CULMs are \textit{Kernel Adaptive Filters} (KAFs) which project the input data into a \textit{Reproducing Kernel Hilbert Space} (RKHS) by using a strictly positive definite kernel function and use linear methods to train parameters \cite{liu2011kernel}. The difficulty is that when employing the representer theorem in RKHS, the filter output is computed from all past data, so the filter computation grows linearly or super-linearly with respect to the sample number $ n $, which is unrealistic for real word applications without sparseness procedures \cite{Liu2008}. The other class, including reservoir computing, uses stochastic approaches based on random hidden parameters exemplified by the \textit{Extreme Learning Machine} (ELM), and it suffers from incomplete theoretic understanding and requires many “tricks” to achieve useful and reproducible results \cite{huang2006extreme}. 
\begin{figure*}[htbp]
	\centering{\includegraphics[width=120mm]{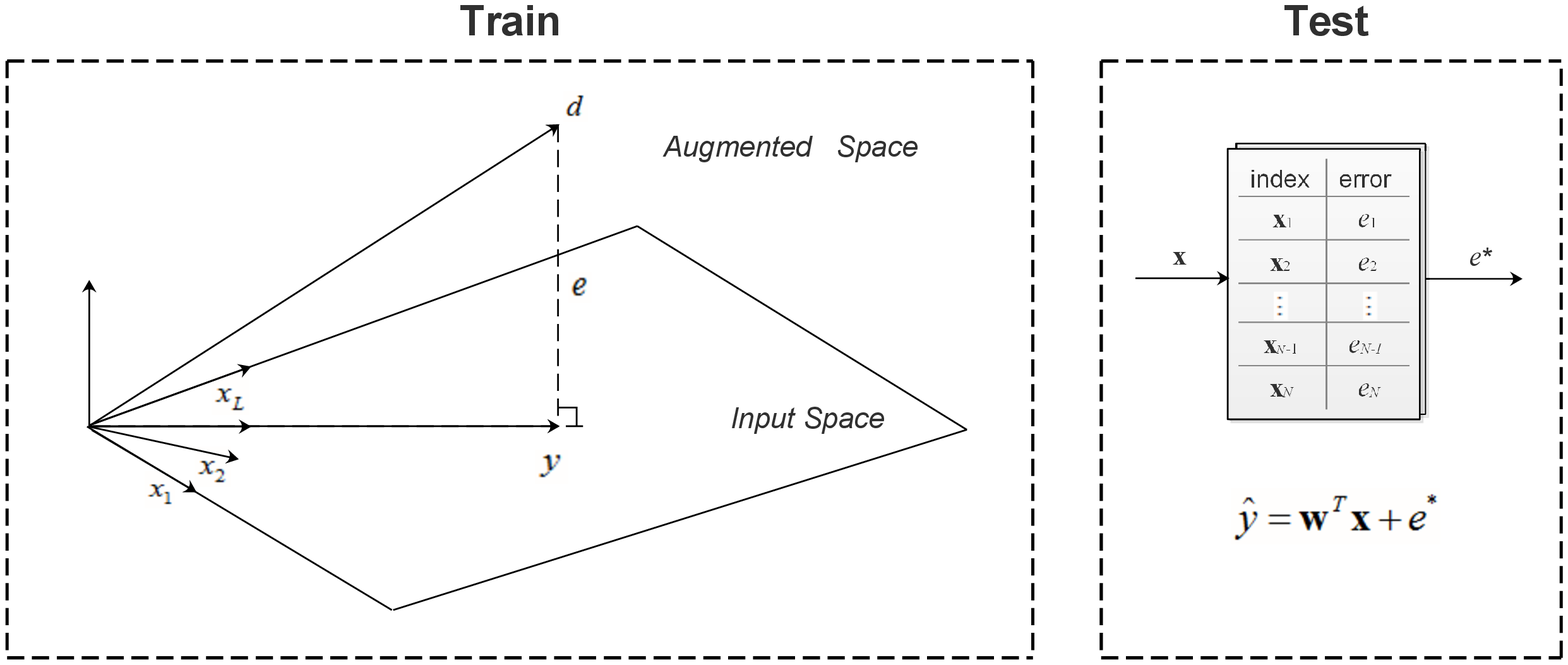}}
	\vspace{-3mm}
	\caption{The application and the geometric structure of ASLM}\label{JS}
	\vspace{-3mm}
\end{figure*}
In spite of these shortcomings, ELMs are surprisingly very popular, which means that the need to achieve fast universal processing with generalization capacity in function approximation is still unmet. 

Here we propose a new solution to design CULMs based on the conventional \textit{Finite Impulse Response} (FIR) linear model extended with a table lookup. Instead of using the input to span the projection space, we use the full joint space as the projection space, hence this approach is named \textit{Augmented Space Linear Model} (ASLM). Augmented with the desired signal, the framework of ASLM expends the data input space, assumed of dimension $ L $, to $ L+1 $ dimensional space. Then the $ L+1 $ independent bases can span any $ L+1 $ space, which means the training set error can be as small as the adaptation method can achieve using a linear approach.

There are two difficulties that need to be addressed in the ASLM. The first is that all the weights go to zero after training except the one that is connected to the desired,  which approaches one. This means we need regularization during the adaptation process. The second difficulty is we don't have the desired signal in the test phase! To address these issues here, we use the difference between the outputs in the input space and the desired in the joint space during the training phase (the training error) to augment the input space instead of expanding the input by the desired. Then, we store all the training errors in a table indexed by the input data. Our novel solution takes advantage of the extra information contained in training errors, which are wasted in conventional least squares, to approach nonlinear relationships with a linear model and a table. Since ASLM is an adaptive linear architecture with convex optimization, and the training error is orthogonal to the bases, the adaptation process no longer needs to be regularized. Meanwhile, the computational complexity of training and testing is much lower than nonlinear methods, which is well suited for online learning algorithms. As a matter of fact, ASLM is an intermediate solution in the complexity-accuracy design space between the linear model (easy but not very accurate) and fully nonlinear models (complex but can be much more accurate). Different from the traditional linear and nonlinear models, the augmented space model framework makes full use of training errors, which may improve others models (linear and nonlinear) as well. 

\section{Augmented Space Linear Model}
The simplest implementation of the ASLM is presented below. The left part of Fig.\ref*{JS} shows the conceptual least square solution of finding the best approximation ($ y $) of the desired response ($ d $) by projecting $ d $ in the space spanned by the multidimensional input $ \textbf{x} $. The minimum error is achieved when $ y $ is the orthogonal projection of $ d $ in the input subspace. It is sufficient to add the error $ e $ to the output $ y $ to obtain exactly the desired response $ d $ in the training set, because it is by definition perpendicular to the input space. When computing the output of the ASLM, obviously, we are using the joint space to evaluate the desired.

Consider a set of N pairs of training input vectors with desired output $ \left\lbrace\textbf{x}_{i},d_{i}\right\rbrace ^{N}_{i=1} $, where $ i $ denotes discrete time instant. We first compute the weights of the linear model in the input space with all the training data, which can be evaluated by the \textit{Least Squares} (LS) solution in (\ref{winner})
\begin{align}\label{winner}
{\bf{w}} = {\left( {\delta {\bf{I}} + {{\bf{X}}^T}{\bf{X}}} \right)^{ - 1}}{{\bf{X}}^T}{\bf{d}}
\end{align}
where $ \textbf{X}=[\textbf{x}_{1},\cdots,\textbf{x}_{N}]^{T} $, $ \textbf{d}=[d_{1},\cdots,d_{N}] $ and $ \delta $ is a small value to prevent rank deficiency. Then we create a table addressed by the input which relates the input with the training error, and store this table. The size of this table will be the training set size if no quantization is introduced. In the test phase, we use the current input to find the closest entry in the table, and then read back the corresponding training set error to approximate the desired response, i.e. equation (\ref{est})
\begin{align}\label{est}
\hat y_{i} = y_{i} + e^* = \textbf{w}^{T}\textbf{x}_{i}+e^*
\end{align}
where $ y_{i} $ is the current output of LS solution, and $ e^* $ is the error obtained from the training set, which is corresponding to the closest $ \textbf{x}_{i} $ in index as we show in right part of Fig.\ref{JS}. The $ l_{2} $ norm is used to measure the distance of transformed samples produced by Hadamard product $ \textbf{w}\!\circ\! \textbf{x} $. Considering the inputs of the whole training set, this Hadamard metric can get more reasonable results when finding the closest sample. The error $ e^* $ in equation (\ref{est}) will be a good approximation for the desired under two conditions: (1): $ e^* $ is a good approximation for the current test sample $ \textbf{x}_{i} $ from the training error; (2): the error in the test set for a given input remains stationary from training and testing.  The second assumption also must be imposed in conventional functional approximation, although in ASLM, the requirement applies to instantaneous errors which is more demanding in practical noisy conditions. In realistic application cases, we can use a quantization approach to cut the noise in the training data while also decreasing the table size.

ASLM is the simplest model in the augmented space, actually, we can also use it to augment the KAF nonlinear model. Although KAFs are universal nonlinear models, it is difficult to achieve a good approximation to the desired by linear combination of Gaussian kernel, because of the rattling and insufficient training data. In order to compute the training error in the augmented space, we first train a nonlinear model as usual, fixed the weights, and compute the training errors all at once. Then we create a table addressed by the input and store this table as mentioned before. In the test phase, we use the current input to find the closest entry in the table, and then read back the corresponding training error to approximate the desired response, i.e. equation (\ref{est}). Since we don't have $ \textbf{w} $ in the nonlinear models, we measure the distance between input samples by $ l_{2} $ norm directly.

In order to improve the efficiency of finding the nearest neighbor, we use a $ kd $ tree to store the data in the table with searching complexity of $ O(\log(N)) $ \cite{samet1990design}. Hence, the testing computational complexity in the augmented space model consists of 2 parts. One is the complexity of the algorithm (linear or nonlinear) to compute the system output, and the other is the complexity of searching for the best error of the augmented space model. The testing computational complexity of ASLM, for example, is $ O(L+\log(N)) $. As for the training, ASLM is very fast, since it only needs to create the table after the least squares algorithm, which is much faster than training a nonlinear model. We will compare the performance and computational complexity of the proposed ASLM with several linear or nonlinear models in the next section.
\vspace{-3mm}
\section{Simulation Results}
\subsection{Prediction Without Noise}
In order to evaluate the role of ASLM within the current methodologies for function approximation and system identification, we select three competing models: the \textit{Least Squares} (LS) as an example of the optimal linear projection, the \textit{Nearest Neighbor} (KNN) algorithm \cite{Cover1967Nearest} as a memory based approach and the \textit{Kernel Least Mean Square} (KLMS), using a Gaussian kernel, as a CULM that rivals in performance with the best nonlinear networks for prediction. We also include KLMS with Augmented space model (KLMS-AM), as an extra comparison, to show the general capabilities of the Augmented space model. All the hyper parameters were validated to get the best possible results including the kernel size $ \sigma $, the step size $ \eta $ and the regularization factor $ \delta $.  For simplicity, we choose $ K=1 $ for the KNN algorithm and all the parameters are showed in the last column of table \ref{noNoise}. The inclusion of the memory based method is judged important because ASLM also uses a table lookup that is similar in spirit to the memory based approaches for modeling. The problem we selected is the prediction of the of the $ x $ component of the Lorenz system \cite{liu2011kernel}, which has been well studied in the literature (order L=7 according to Takens’ embedding theorem) \cite{lorenz1963deterministic}. The Lorenz data set is generated from the differential equation with the parameters $ \beta={\raise0.7ex\hbox{$8$} \!\mathord{\left/	{\vphantom {8 3}}\right.\kern-\nulldelimiterspace}\!\lower0.7ex\hbox{$3$}}, \delta=1,\rho=28 $. A first order approximation is used with a step size parameter $ 0.01 $. Segments of $ 2000 $ samples are used as the train set and the following $ 400 $ samples are the testing set. We normalize the time series to zero-mean with unit variance. Performance is measured as the power of the error. Results are averaged over $ 50 $ independent training-test runs obtained by sliding the window over the generated data by $ 50 $ samples each time. 
\vspace{-2mm}
\begin{table}[h] \footnotesize 
	\caption{\label{noNoise}The performance comparison of linear and nonlinear algorithm} 
	\begin{tabular}{|C{1.5cm}<{\centering}|p{3cm}<{\centering}|p{1cm}<{\centering}|p{1.4cm}<{\centering}|} 
		\hline  
		Algorithm & Testing MSE & Training MSE & Parameter\\ 
		\hline 
		\hline 
	    KLSM-AM & $\!\!\!\!5.71\! \times\! {10^{ \!-\! 4}} \pm 2.83\! \times \!{10^{ \!-\! 4}}\!\!\!\! $ & $ 0 $ & $ K=1 $\\  
	    \hline  
	    KLMS & $\!\!\!\!2.74\! \times\! {10^{ \!-\! 3}} \pm 8.43\! \times \!{10^{ \!-\! 4}}\!\!\!\! $ & $ \!\!\!\!2.22\!\! \times\!\! {10^{ \!-\! 3}} \!\!\!\!\!$ & $\!\!\!\! \sigma\!=\!1\!,\eta\!=\!0.7\!\!\!\! $\\  
		\hline  
		ASLM & $\!\!\!\!3.13\! \times\! {10^{ \!-\! 3}} \pm 6.97\! \times \!{10^{ \!-\! 4}}\!\!\!\! $ & $ 0 $ & $ K=1 $\\  
		\hline  
		KNN & $\!\!\!\!1.02\! \times\! {10^{ \!-\! 2}} \pm 2.79\! \times \!{10^{ \!-\!3}}\!\!\!\! $ & $ 0 $ & $ K=1 $\\ 
		\hline  
		LS & $\!\!\!\!2.64\! \times\! {10^{ \!-\! 1}} \pm 1.20\! \times \!{10^{ \!-\! 2}}\!\!\!\! $ & $ \!\!\!\!2.65\!\! \times\!\! {10^{ \!-\! 1}} \!\!\!\!\!$ & $ \delta=0.1 $\\
		\hline   
	\end{tabular}  
\end{table} 
\vspace{-2mm}

We also show the testing MSE and training MSE in table \ref{noNoise}. Since KLMS is an online algorithm while the other algorithms are batch based, the testing MSE of KLMS is calculated from the last 100 points of the converged learning curve. In terms of performance, we see that the LS is the worst performer. Even KNN is better, but we notice that ASLM always improves KNN performance for the same storage. KLMS is a little better than ASLM, which can be further improved by the augmented space model as well. However, when we take into consideration accuracy and computation time, ASLM appears as a very good compromise between the performance of the nonlinear and the linear model. 
\begin{figure}[htbp]
	\centering{\includegraphics[width=80mm]{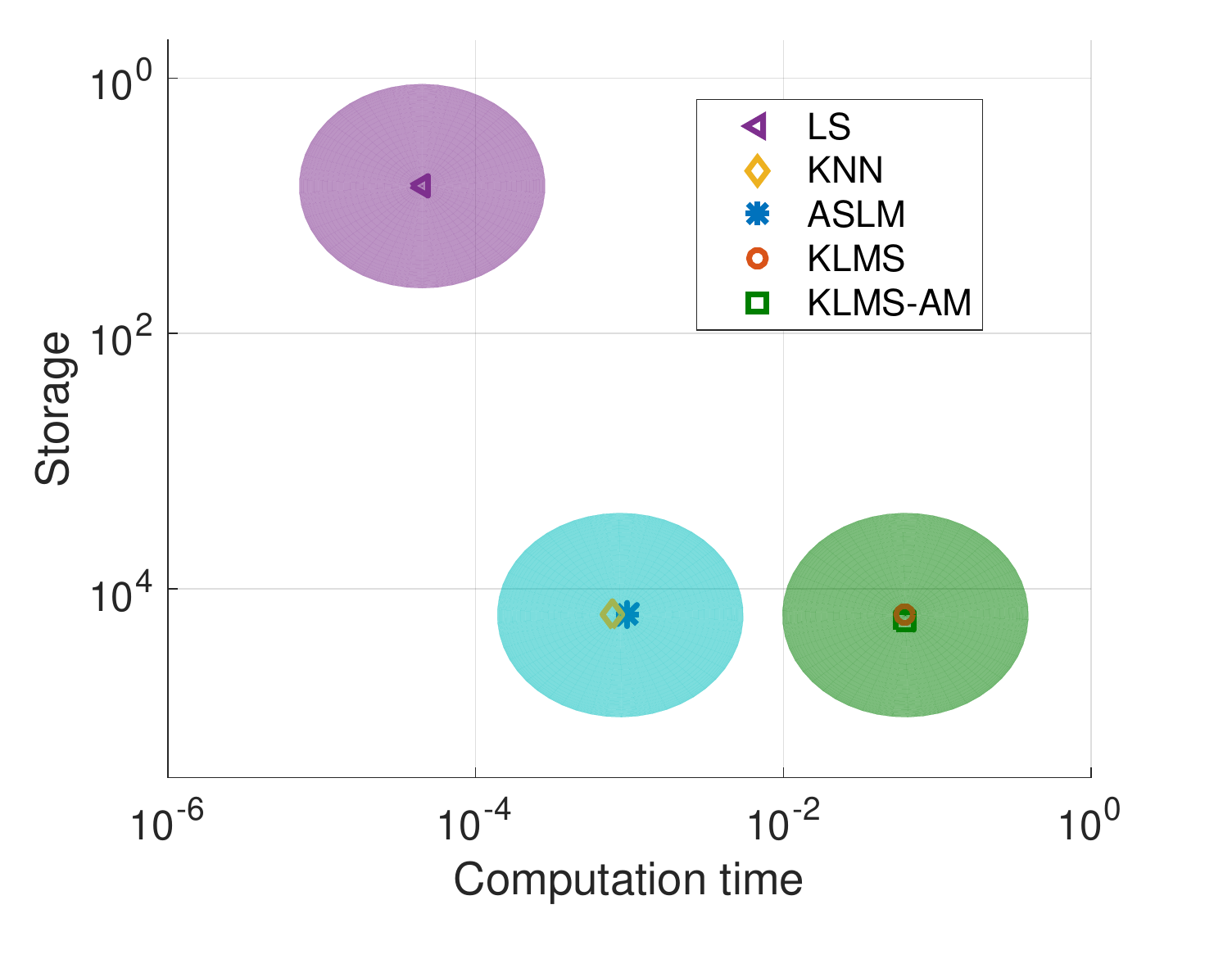}}
	\vspace{-9mm}
	\caption{The computation time and storage comparison of linear and nonlinear algorithm }\label{area}
	\vspace{-2mm}
\end{figure}
Fig. \ref{area} shows the computation time and storage in the test phase of the compared algorithms. In terms of simplicity of resources, the LS solution is unbeatable both in terms of storage and computation time. Compared with KLMS and KLMS-AM, ASLM is much faster with comparable performance, which is much better than the performance of LS algorithm. The bottleneck of ASLM is the search for the best candidate in the table look up, which is very similar to KNN. The general locations of linear and nonlinear model with respect to storage and computational time in this simulation are plotted as an ellipsoid cloud around points.
It is obvious that ASLM is a linear model with nonlinear regression capacity, and its location in the Fig. 2 deviates from the diagonal linking the two linear and nonlinear models, which shows its efficiency. In this simulation, the augmented space model also shows surprising potential to improve different models, which can make full use of the training error or desired in the augmented space to increase the performance. Meanwhile, the computational complexity of searching in the table is much smaller than that of KLMS, so the performance improvement won't bring a big computational burden and explains Fig. 2.
\vspace{-1mm}
\subsection{Prediction With Noise}
In this section, the data are the same as the last experiment and the desired signal of training data are corrupted by 20 dB zero mean Gaussian noise. Performance is measured again as the power of the error. Results are averaged over $ 50 $ independent training-test runs obtained by sliding the window over the generated data by $ 50 $ samples each time. The purpose of the experiment is to compare the performance when the training set is not clean. It is obvious that the performance of ASLM will suffer when the noise is added in the table. Therefore, we use a simple (sequential) \textit{Vector Quantization} (VQ) method to cut the noise,  which can build a small size codebook with a small threshold $ \epsilon $ instead of the original input values in the table. Depending on the Euclidean distance, VQ is computationally simple (linear complexity in terms of the codebook size) \cite{chen2012quantized,Liu2008}. The error of one center in the codebook is computed by averaging the training errors whose indexes are quantized into the same center. This method is first used in KLMS to constrain the network size. In ASLM, VQ not only decreases the table size and computational complexity, but also improves the performance by averaging the local training errors. Three extra comparisons are added to show the improvements of algorithms brought by VQ, which is QKLMS, KLMS-QAM and QASLM. To be fair, the final size of codebooks are set to 500, and all the hyper parameters are validated to get the best possible results (kernel size $ \sigma $, step size $ \eta $, quantization radium $ \epsilon $ and regularization factor $ \delta $). 

\vspace{0mm}
\begin{table}[h] \footnotesize 
	\vspace{-3mm}
	\caption{\label{Noise}The performance comparison of linear and nonlinear algorithm} 
	\newcommand{\tabincell}[2]{\begin{tabular}{@{}#1@{}}#2\end{tabular}}
	\begin{tabular}{|C{1.6cm}|p{2.7cm}<{\centering}|R{1.1cm}<{\centering}|p{1.5cm}<{\centering}|} 
		\hline   
		Algorithm & Testing MSE & \tabincell{c}{Training \\ MSE} & Parameter\\ 
		\hline 
		\hline
		KLMS & $\!\!\!8.24\! \times\! {10^{ \!-\! 3}} \pm 1.66\! \times \!{10^{ \!-\! 2}}\!\!\!\! $ &  $ \!\!\!\!1.65\!\! \times\!\! {10^{ \!-\! 3}} \!\!\!\!\!$  & $\!\!\!\! \sigma\!=\!1\!,\eta\!=\!0.7\!\!\!\! $\\  
		\hline 
		QKLMS & $\!\!\!8.28\! \times\! {10^{ \!-\! 3}} \pm 1.26\! \times \!{10^{ \!-\! 3}}\!\!\!\! $ & $ \!\!\!\!1.70\!\! \times\!\! {10^{ \!-\! 2}} \!\!\!\!\!$ & \tabincell{c}{$\!\!\!\! \sigma\!=\!1\!,\eta\!=\!0.7\!\!\!\!$\\ $\epsilon=0.085$}\\  
		\hline  
		KLSM-AM & $\!\!\!1.04\! \times\! {10^{ \!-\! 2}} \pm 7.27\! \times \!{10^{ \!-\! 4}}\!\!\!\! $ & $ 0 $ & \tabincell{c}{$\!\!\!\! \sigma\!=\!1\!,\eta\!=\!0.7\!\!\!\!$\\ $K=1$}\\  
		\hline 
		KLSM-QAM & $\!\!\!4.42\! \times\! {10^{ \!-\! 3}} \pm 6.48\! \times \!{10^{ \!-\! 4}}\!\!\!\! $ & $ \!\!\!\!8.42\!\! \times\!\! {10^{ \!-\! 3}} \!\!\!\!\! $ & \tabincell{c}{$\!\!\!\! \sigma\!=\!1\!,\eta\!=\!0.7\!\!\!\!$\\$\!\!\!\!  K\!\!\! =\!\!1\!,\!\epsilon\!\!=\!0.085\!\!\!\! $}\\  
		\hline  
		ASLM & $\!\!\!1.32\! \times\! {10^{ \!-\! 2}} \pm 1.25\! \times \!{10^{ \!-\! 3}}\!\!\!\! $ & $ 0 $ & $ K=1 $\\  
		\hline 
		QASLM & $\!\!\!1.03\! \times\! {10^{ \!-\! 2}} \pm 1.20\! \times \!{10^{ \!-\! 3}}\!\!\!\! $ & $  \!\!\!\!1.10\!\! \times\!\! {10^{ \!-\! 2}} \!\!\!\!\! $ & $\!\!\!\! K\!\!\!=\!\!1\!,\!\epsilon\!\!=\!0.032\!\!\!\! $\\  
		\hline  
		KNN & $\!\!\!2.02\! \times\! {10^{ \!-\! 2}} \pm 8.85\! \times \!{10^{ \!-\!4}}\!\!\!\! $ & $ 0 $ & $ K=1 $\\ 
		\hline  
		LS & $\!\!\!2.64\! \times\! {10^{ \!-\! 1}} \pm 1.52\! \times \!{10^{ \!-\! 2}}\!\!\!\! $ & $ \!\!\!\!2.74\!\! \times\!\! {10^{ \!-\! 1}} \!\!\!\!\!$ & $ \delta=0.1 $\\
		\hline   
	\end{tabular}  
\vspace{-4mm}
\end{table} 
We show the testing MSE, training MSE and the corresponding parameters in the table \ref{Noise}. It is easy to notice that the linear model shows powerful robustness in this comparison, since the performance of all the algorithms except LS are affected compared with the last result. ASLM can beat KNN and both algorithms get better results than LS algorithm. Without the quantization method, KLMS is the best predictor in this experiment, which is more robust than KLMS-AM and ASLM, because the sum of weighted Gaussian kernel can remove the noise to some extent. VQ reduces the storage for KLMS, KLMS-AM and ASLM algorithms, while it improves the performance of KLMS-AM and ASLM by cutting the noise in the table. However, noise removal by of VQ is not obvious for KLMS algorithm. Hence, KLMS-QAM shows the best performance with the help of VQ. QKLMS shows similar result with KLMS, which is still better than QASLM.
\vspace{-1.5mm}
\section{Conclusion And Discussion}
We presented a new solution to the functional approximation problem, by taking advantage of the linear solution, and correcting this estimate with the training error from the input sample that is in the neighborhood of the current test phase input. In essence, we combine the computational efficiency of the linear solution with a memory block encoding the training errors originating from the nonlinearity of the data generation process, which produce a nonlinear response. In conventional nonlinear function approximation, one needs to find appropriate parameters of nonlinear mappers, which is full of difficulties and also computationally expensive. This is the reason ASLM displays an interesting compromise in the space of accuracy and computation complexity between the conventional linear and nonlinear solutions. However, noticing that the ASLM only models the linear error in this case, we have also shown that the same approach can improve upon the modeling of the nonlinear error with the KLMS-AM.  

ASLM is a member of the CULM family that has not been investigated in the past. Conceptually, we are proposing to augment the input projection space with the desired response, so this opens the door to study many different implementations of the simple ASLM discussed in this paper. The thrust of research should focus on ways to improve the table look up performance which is very rudimentary. In noisy situations, we can think of PCA to obtain a better definition of the input space, and filter the training error by local modeling. In fact, It is very interesting to interpret the training errors as a sensitivity to the unknown desired response that can be exploited for Bayesian modeling. Since we have an implicit model of the input, we can also speed up the search to find the closest neighbor of the current input. These simple modifications will improve ASLM performance and lead to new applications beyond functional approximation. We therefore believe that this will be a vibrant line of research for years to come.

\vfill\pagebreak

\bibliographystyle{IEEEbib}
\bibliography{my}

\end{document}